\pdfoutput=1

\documentclass[11pt,table]{article}

\usepackage[preprint]{acl}
\usepackage{times}
\usepackage{latexsym}
\usepackage[T1]{fontenc}
\usepackage[utf8]{inputenc}
\usepackage{microtype}
\usepackage{inconsolata}
\usepackage{graphicx}
\usepackage{listings}
\usepackage{ulem}
\usepackage{amsmath}
\usepackage{multirow}
\usepackage{subcaption}
\usepackage{hyperref}       
\usepackage{url}            
\usepackage{booktabs}       
\usepackage{amsfonts,amssymb}       
\usepackage{nicefrac}       
\usepackage{microtype}      
\usepackage{xcolor}         
\usepackage{ulem}
\usepackage{soul}
\usepackage{xcolor}

\usepackage[utf8]{inputenc}
\usepackage{xspace}
\newcommand*{\eg}{e.g.\@\xspace}
\newcommand*{\ie}{i.e.\@\xspace}

\newcommand{\cmmnt}[1]{\ignorespaces}

\title{FactSim: Fact-Checking for Opinion Summarization}

\author{Leandro Anghinoni \\
  MercadoLibre \\
  São Paulo, Brazil \\
  \texttt{leandro.anghinoni@mercadolivre.com} \\\And
  Jorge Sanchez \\
  MercadoLibre \\
  Cordoba, Argentina \\
  \texttt{jorge.sanchez@mercadolibre.com} \\}

\begin{document}
\maketitle
\begin{abstract}
We explore the need for more comprehensive and precise evaluation techniques for generative artificial intelligence (GenAI) in text summarization tasks, specifically in the area of opinion summarization. Traditional methods, which leverage automated metrics to compare machine-generated summaries from a collection of opinion pieces, \eg product reviews, have shown limitations due to the paradigm shift introduced by large language models (LLM). This paper addresses these shortcomings by proposing a novel, fully automated methodology for assessing the factual consistency of such summaries. The method is based on measuring the similarity between the claims in a given summary with those from the original reviews, measuring the coverage and consistency of the generated summary. To do so, we rely on a simple approach to extract factual assessment from texts that we then compare and summarize in a suitable score. We demonstrate that the proposed metric attributes higher scores to similar claims, regardless of whether the claim is negated, paraphrased, or expanded, and that the score has a high correlation to human judgment when compared to state-of-the-art metrics.
\end{abstract}

\section{Introduction}

\begin{figure*}[ht]
\footnotesize
\centering
\begin{tabular}{ p{2.0cm} p{13.0cm} }
\hline
\\
Claims (frequency) & \textcolor{blue}{Nice/Great (6)}, \textcolor{teal}{Durable (5)}, \textcolor{red}{Good Fit (5)}, \textcolor{orange}{Convertible (3)}, \textcolor{purple}{Soft (2)}, \textcolor{cyan}{Comfortable (2)}, \textcolor{magenta}{Runs Small (2)}, Perfect (1), Well Made (1), Nice Color (1), Pretty (1)\\
\\
\hline
\\
Reference summaries & [\textcolor{blue}{These tights are great}. \textcolor{teal}{They are durable and do not tear easily, they can be worn and washed without worry}. \textcolor{orange}{The bottoms of can be pulled up easily so that sandals can be worn with them}. \textcolor{magenta}{It might be a good idea to order a size bigger because they can be a little tight in the waist.} Overall, these tights are definitely recommended.] \\

 & [\textcolor{cyan}{These transition tights are perfect for children sensitive to the tight sensation other tights have around the foot}.  \textcolor{purple}{The material is soft} and \textcolor{teal}{durable; they stand up well to both the rough nature of children, and the washing machine.}  \textcolor{magenta}{This product does tend to run slightly small, so purchasing one size up is recommended.}] \\

& [Bought these for my 3-year-old daughter for her classes and \textcolor{blue}{they turned out great} and \textcolor{red}{fit perfectly}. \textcolor{orange}{She can pull them up to walk in then pull down to cover her toes for class}. \textcolor{teal}{Strong tights that should last}. \textcolor{magenta}{You might need to buy a size up just for comfort around the waist.}] 
\\
\\
\hline
\end{tabular}
\caption{Reference summaries for product B000A2FTN6 evaluation data of proposed by \citet{bravzinskas2019unsupervised}. In the first row, we show the claim frequency in the original eight reviews for the same product. Notice that none of the three reference summaries is able to cover the colored claims (frequency > 1), which should be more relevant. Also, the wording favors extractive models rather than abstractive ones, especially in metrics such as ROUGE and BERTScore.}

\label{fig:reference_summ}
\end{figure*}

The recent advances in generative artificial intelligence (GenAI) have enabled foundation models to achieve robust results in a variety of tasks, such as text summarization \cite{achiam2023gpt}, sentiment analysis, translation and question answering \cite{chang2024survey}. Innovations in deep learning and advances in computational power have enabled these models to process content at a deeper level, providing more comprehensive summarizations.

Summarization tasks find utility in various domains, each with unique requirements. Most of them, such as news summarization, aim to extract and compile key points from one lengthier source document to create a concise summary. This emphasizes the adherence of facts, such as events, places, and dates, to information provided by a source document without a guarantee that the information is true \cite{kryscinski2019evaluating}. On the other hand, traditional opinion summarization seeks to process a collection of smaller texts, such as reviews about a particular product or service and explores subjective aspects, processing the emotions, sentiments, and stance of a person \cite{shen2023opinsummeval}. With the exponential growth of online reviews and comments, generating a single summary review that contains all the facts with the right sentiment from all the source reviews is extremely valuable, but also challenging since it involves weighting claims from multiple documents \cite{kim2022beyond}. 

Evaluation of summarization methods usually involves comparing the machine-generated summaries to reference summaries, typically curated by humans \cite{shen2023opinsummeval}. In this process, automated metrics are commonly used, such as BLEU \cite{papineni2002bleu} and ROUGE \cite{lin2004rouge}, which mostly check n-gram overlap, and BERTScore \cite{zhang2019bertscore}, which is better at capturing semantic equivalence. Despite their prevalent use, recent works have cast doubt on these evaluation protocols, especially with the paradigm shift brought by large language models (LLM). Some works suggest that recent models such as gpt-4 \cite{achiam2023gpt}, perform worse than older models like BRIO \cite{liu2022brio} in traditional automated metrics, despite the human preference for their summaries \cite{shen2023opinsummeval}. Other works state that current models can surpass reference summaries crafted by humans in quality, which puts in question whether the current process can overlook factual inconsistencies stemming from the wrong interpretation of data \cite{goyal2022news}. In Fig.~\ref{fig:reference_summ} we provide a qualitative example, where we show that none of the three reference summaries is able to entail all the claims in the original reviews and that some of them do not include the most relevant considerations about the product.

Large language models present high abstractive capabilities, being able to comprise the original information in a completely rephrased text. This is a challenge for n-gram matching (as some words may have been replaced by synonyms) and for sentence similarity schemes (as the facts from different sentences may have been comprised in a single sentence). Due to these issues, neural-based schemes such as BARTScore \cite{yuan2021bartscore} and self-checking schemes have flourished \cite{manakul2023selfcheckgpt}. In neural-based schemes a pre-trained backbone model is used to evaluate the probability that a summary was generated by a certain source text, while in self-checking models the summary entailment is validated by the LLM itself, which asks structured questions about the summary in a reference-free manner. These methods, however, lack explainability and are prone to inconsistencies inherited from the pre-trained backbone model.

Another limitation of current evaluation methods is that different domains present different summarization objectives. Summarizing a set of smaller documents, like opinions, differs inherently from summarizing a single large document as maintaining factual consistency in opinion summaries relates more to the consensus among various sources. Also, evaluating the factual accuracy of large text summaries may need access to external sources to check if the claims are true or not, whereas the factual consistency of opinion summarization is restricted to cover the most cited claims among the source texts, whether they are true or not. Therefore, a need arises for more comprehensive and robust evaluation techniques that can ponder over the statistical properties of the claims.

In this work, we propose a novel protocol to evaluate the factual consistency of summaries, specifically for the summarization of product reviews for e-commerce. Our work proposes the atomization of sentences into fact tuples that can be compared semantically to generate scores in terms of coverage and consistency of the summary.

\begin{figure*}[ht]
  \centering
  \includegraphics[width=1.0\linewidth]{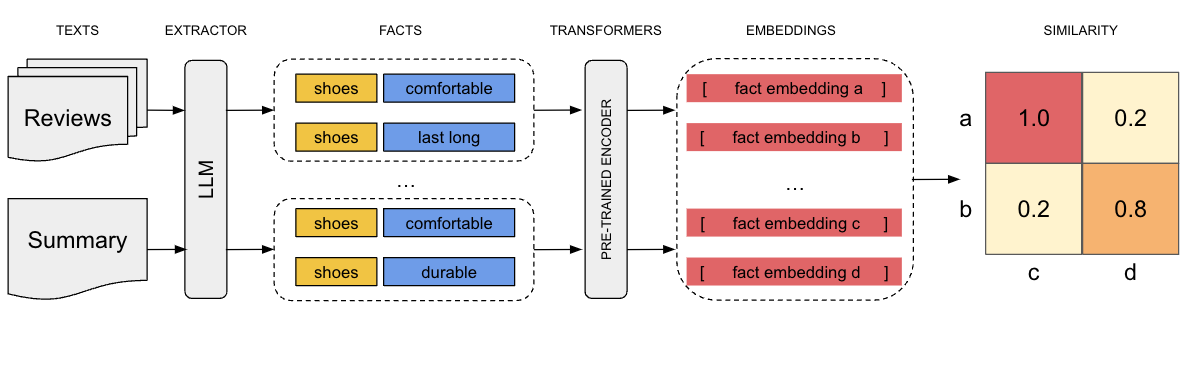}
  \caption{Overview of the proposed method. First, we extract fact tuples from both the original reviews and the summary. Then we project the tuples in a semantic embedding space using a pre-trained encoder. Finally, tuple embeddings are compared using the cosine similarity and summarized into a final metric.}
  \label{fig:overview}
\end{figure*}

The contributions of this work are as follows:

\begin{itemize}
    \item We propose a text fact extractor based on LLM prompt engineering that handles negation, paraphrasing and text expansion; 
    \item We propose a new metric ($FactSim$) to evaluate the fact coverage and consistency of the summary; 
    \item The approach is automatic like traditional metrics, but reference-free like self-checking schemes;
    \item We found that our metric has a high correlation to human judgments and is better at scoring paraphrased claims than n-gram overlap methods while being, at the same time,  explainable.
\end{itemize} 

This work is organized as follows: In Section \ref{related-work} we give a brief overview of the current state of opinion summarization evaluation. In Section \ref{method} we present our method and detail its differences from related frameworks. Next, in Section \ref{experiments} we detail the datasets and the setups for the experiments. In Section \ref{sec:results} we present empirical results and benchmark comparison of the proposed metric. Finally, in Section \ref{conclusion} we give our final remarks.

\section{Opinion Summarization Evaluation}
\label{related-work}

Human evaluation protocols \cite{kim2022beyond} are considered the gold standard for evaluating opinion summaries. Despite being time-consuming and potentially resource-intensive, these protocols enable a more comprehensive rating of the summaries because they focus not only on accuracy but also on qualitative dimensions. Human assessors can evaluate aspects like coherence, relevance, and readability that automated metrics may overlook. Although human judgment is still prone to error, automated metrics usually seek a high correlation with the human score.

Automated metrics, such as ROUGE \cite{lin2004rouge}, BLEU \cite{papineni2002bleu} and BERTScore \cite{zhang2019bertscore}, have become pivotal tools in the evaluation of summarization tasks. ROUGE, a statistically-based method, is widely employed due to its reliability in content overlap calculation between the generated and reference summaries, aiding in the evaluation of a summarization system's performance. Meanwhile, BERTScore offers a context-focused algorithm catering to the fine specificities of text and comprehension nuances. Its innovation stems from its application of pre-trained language models to compute similarity scores on token-level embeddings, providing a more context-sensitive summary. It is the precursor of several neural-based scores, such as BARTScore \cite{yuan2021bartscore} and BLANC \cite{vasilyev2020fill}.

QA-based models such as QAFactEval \cite{fabbri2021qafacteval} and MQAG \cite{manakul2023mqag} are effective in determining how well a summarization system performs based on pre-defined questions that aim to reproduce equal answers from both source and summary texts. One major utility of QA is its ability to break down the information into simpler questions that can be evaluated in a binary fashion. However, how an evaluative QA system can adequately cope with the opinionated aspect of summaries is still a matter of exploration.

More recently, GPT-based models have paved the way for self-evaluation mechanisms \cite{manakul2023selfcheckgpt} in the domain of opinion summarization. These models, trained on massive text corpora, have shown promising results in being able to generate summaries and then assess their own performance. Yet, self-evaluation protocols struggle with trustworthiness and explainability since one is evaluating a black-box model with a black-box evaluator. In light of that, mixed models have been proposed, where important information is extracted from both the source and the summary and then compared using similarity metrics. Some works have proposed reducing the source document to a set of relevant sentences \cite{laban2022summac}, others have used a smaller granularity by retrieving atomic facts from the text \cite{hu-etal-2023-refchecker, min2023factscore,goodrich2019assessing}

Although these mixed models have achieved impressive results, some problems remain an open question, especially concerning opinion summarization. First, traditional models are evaluated against a reference summary, generated by a human, which we argue might not be the best choice in the case of opinion summarization. Second, statistical properties of the source document are neglected in these models, since they are usually tailored to the task of reducing one lengthier text into a summary. The model we propose is inspired by some of these recent models and aims to overcome the aforementioned limitations.

\section{{F\MakeLowercase{act}S\MakeLowercase{im}}: A new method to evaluate text summarization}
\label{method}

We aim to measure the quality of a summary generated from a possibly large set of reviews, each of which focuses on either positive or negative aspects (or both) of a subject or topic. The content of these reviews is not required to be exhaustive with respect to the properties being described, as they arise from the subjective considerations of their authors. Our goal is to provide a metric that quantifies the quality of a summary in terms of relevance and factual consistency with respect to the information agreed upon as the most relevant given the pool of input reviews.

An overview of our method is outlined in Fig.~\ref{fig:overview}. We start by extracting a set of "facts tuples" independently from both the summary and the pool of reviews given as inputs. These tuples aim at encoding factual assertions about the objects being reviewed. We force them to be concise and short by explicitly conditioning the prompt to comply with such premises. Tuples are embedded into a vector space using a pre-trained encoder. We use the similarity between these embeddings to derive a metric. In what follows, we provide details of each step of the process.

\subsection{Formulation}

Let $R=\{r_1, r_2, \dots, r_K\}$ be the set of source text reviews, and $s$ the candidate summary. We compute a set of fact tuples from the summary and each review in $R$. Fact tuples are objects $f=(o,d)$, where $o$ denotes a property or element of the subject being described and $d$ its quality, \eg a sentence as "the shoes are comfortable" is encoded as \textit{(shoes, comfortable)}. A tuple can repeat multiple times among the different reviews in $R$. Also, the use of correferences, synonyms, and negations might result in syntactically different but semantically similar tuples. The tuple extraction method is key for capturing such variations. We can leverage the paraphrasing capabilities of current LLMs to deal with this problem, as described later in this section.
We denote as $F^S=\{f^S_1, f^S_2, \dots, f^S_M\}$ and $F^R=\{f^R_1, f^R_2, \dots, f^R_N\}$ the fact tuples extracted from the summary and reviews, respectively. The latter corresponds to the concatenation of tuples extracted from reviews in $R$. Let $v=g(f)\in\mathbb{R}^d$ be a $d$-dimensional vector representation of tuple $f$ computed by an encoder $g$. We define the following scores:
\begin{equation}
    f_V(F^R, F^S) = \frac{1}{N} \sum_i^N \max_{j=1, \dots, M} sim(v^R_i, v^S_j),
    \label{eq:fV}
\end{equation}  
\begin{equation}
    f_N(F^R, F^S) = \frac{1}{M} \sum_j^M \max_{i=1, \dots, N} sim(v^R_i, v^S_j),
    \label{eq:fN}
\end{equation}
where $sim$ denotes a suitable similarity function. 

Eq.~\eqref{eq:fV} (\eqref{eq:fN}) can be seen as the average of similarities between review (summary) tuples and their most similar ones from the summary (reviews). $f_V$ measures the \emph{co\textbf{V}erage} of the summary with respect to the set of facts present in the source documents. A score of $1$ would indicate that every fact in the source reviews is present in the summary, with the exact same choice of words (no paraphrasing). Also, this metric takes into account the relevance of a fact in the source documents, \ie if a fact is mentioned in several reviews it will contribute a higher score to the final metric since that specific description is a consensus among different reviews. However, it does not measure if a summary diverges from the source documents by adding facts that are not present in the input texts, \eg hallucinations. The introduction of Eq.~\eqref{eq:fN} allows us to characterize the \emph{co\textbf{N}sistency} of the facts encoded in the summary with respect to those present in the source reviews. A score of $1$ means that every aspect highlighted in the review exists in at least one of the documents in the input.
Based on the above, we define \textit{{F}act{S}im} as the harmonic mean of the these scores, 
\begin{equation}
\begin{split}
    \text{FactSim}&(F^R, F^S) = \\ 
    & 2\frac{f_V(F^R, F^S) f_N(F^R, F^S)}{f_V(F^R, F^S) + f_N(F^R, F^S)},
\end{split}
\end{equation}
where we assume $sim$ is defined over $[0,1]$. Here, a high value for FactSim indicates high values for $f_V$ and $f_N$.

\subsection{Extraction of fact tuples}
{
\renewcommand{\arraystretch}{1.1}
\begin{table*}
\centering
\small
\begin{tabular}{ll}\toprule
Input sentence &Fact-tuples \\\midrule
1. The car is not good, it is not fast at all &\texttt{[['car', 'bad'], ['car', 'slow']]} \\
2. The car is good, it is very fast &\texttt{[['car', 'good'], ['car', 'fast']]} \\
3. The car is bad, it is slow &\texttt{[['car', 'bad'], ['car', 'slow']]} \\
4. The car is great, it is quick &\texttt{[['car', 'great'], ['car', 'quick']]} \\
\bottomrule
\end{tabular}
\caption{Toy example showing different sentences and extracted tuples.  }
\label{tab:toy_examples}
\end{table*}
}

\begin{figure*}[ht]
\centering
\begin{subfigure}[c]{0.33\textwidth}
  \centering
  \includegraphics[width=0.8\linewidth]{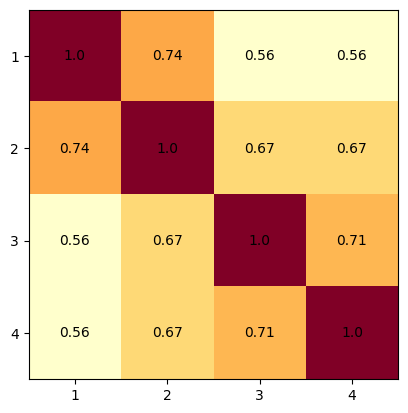}
  \caption{ROUGE-1 (R-1)}
  \label{fig:sfig1}
\end{subfigure}%
\begin{subfigure}[c]{0.33\textwidth}
  \centering
  \includegraphics[width=0.8\linewidth]{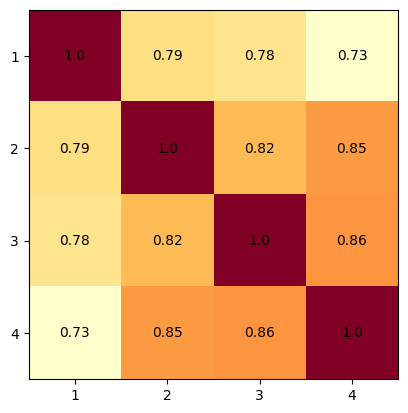}
  \caption{BERTScore (BS)}
  \label{fig:sfig2}
\end{subfigure}%
\begin{subfigure}[c]{0.33\textwidth}
  \centering
  \includegraphics[width=0.8\linewidth]{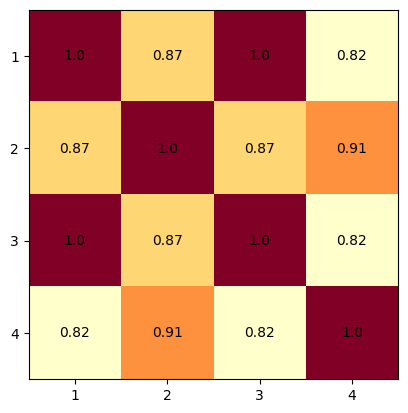}
  \caption{FactSim}
  \label{fig:sfig3}
\end{subfigure}
\caption{Pairwise score matrices for the toy examples in Table~\ref{tab:toy_examples}. 
ROUGE-1 and BERTScore attribute a higher score to pairs 1-2 and 3-4, while the correct pairings 1-3 and 2-4. FactSim gives a consistently higher score to matching pairs than to non-matching ones.}
\label{fig:metrics_comparison_matrices}
\end{figure*}

Extracting structured information from texts is a common task in natural language processing (NLP). While traditional methods build upon identifying patterns in the lexical and syntactic structure of text \cite{han2020more}. Although very efficient in controlled environments, defining rules for every possible pattern is challenging and error-prone. Given the nature of the task, we opt to leverage the abstractive and paraphrasing capabilities of current LLMs via prompt engineering. Relying on high-capacity models has the following advantages:
\begin{itemize}
    \item paraphrasing multi-word expressions
    \item handling negative expressions (\eg rephrase "not good" as "bad")
\end{itemize}
Listing~\ref{lst:fact_tuples} shows the prompt we use for this task. By restricting the LLM to this simpler task allows us to rely on high-capacity models like gpt-4 for processing or to use these models to generate samples for fine-tuning. 

\begin{lstlisting}[language={}, caption=Prompt for fact-tuples extraction, label=lst:fact_tuples]
Your task is to extract knowledge graph tuples from customer reviews and return them as a list. 

Tuples must adhere to the following rules:
1. Tuple are represented as ('subject', 'description')
2. The 'subject' is either the product being reviewed or a feature of the product. It must be one word.
3. The 'description' must be one word, paraphrase it if needed

### Customer reviews:
\end{lstlisting}
Table~\ref{tab:toy_examples} provides some illustrative examples of sentences and the corresponding fact tuples extracted with gpt-4. Note how the model is able to paraphrase 'not fast at all' into 'slow'. This facilitates consistency check as each claim is converted into one description.

\section{Experiments}
\label{experiments}

In the first set of experiments, we use a small toy example to highlight some key properties of the model. We generate a set of synthetic sentences in a way that half of them entail the claim of a reference sentence and the other half contradict it. The sentences were written to include negation, paraphrasing, and text expansion. 

Then, to assess how the proposed metric compares to state-of-the-art, we run experiments using the OpinSummEval \cite{shen2023opinsummeval} benchmark. 

\subsection{Datasets}

\paragraph{OpinSummEval \cite{shen2023opinsummeval}.} OpinSummEval is a dataset built upon Yelp \cite{chu2019meansum}. The dataset contains 100 samples, with the output of 14 summarization models built upon 8 different reviews. Summaries were ranked by humans along four different dimensions, namely: aspect relevance, self-coherence, sentiment consistency, and readability. Aspect relevance measures if the main aspects of the reviews are present in the summary. Self-coherence measures whether the summary does not contain conflicting opinions. Sentiment consistency measures the consistency between aspects and overall sentiment in the reviews. Readability measures whether the summary is fluent and informative. Evaluation is performed by computing the correlation between each automated metric and the human ratings provided with the dataset.

{
\renewcommand{\arraystretch}{1.1}
\begin{table*}[ht]
\footnotesize
\centering
\resizebox{\linewidth}{!}
{
\begin{tabular}{llcccccc}\toprule
Claim &Tuple &Type &FactSim &R-1 &R-2 &R-L &BS \\\midrule
The car is fast* &\scriptsize\texttt{[['car', 'fast']]} & - &1.00 &1.00 &1.00 &1.00 &1.00 \\
The car is not slow &\scriptsize\texttt{[['car', 'fast']]} &Ent. N &1.00 &0.67 &0.57 &0.67 &0.88 \\
The car is quick &\scriptsize\texttt{[['car', 'quick']]} &Ent. P &0.84 &0.75 &0.67 &0.75 &0.93 \\
The car is pretty fast around the corners &\scriptsize\texttt{[['car', 'fast']]} &Ent. L &1.00 &0.67 &0.40 &0.67 &0.72 \\\midrule
The car is slow &\scriptsize\texttt{[['car', 'slow']]} &Contr. P &0.76 &0.75 &0.67 &0.75 &0.94 \\
The car is not fast &\scriptsize\texttt{[['car', 'slow']]} &Contr. N &0.76 &0.89 &0.57 &0.89 &0.90 \\
The car is sluggish &\scriptsize\texttt{[['car', 'sluggish']]} &Contr. P &0.53 &0.75 &0.67 &0.75 &0.76 \\
The car is pretty slow around the corners &\scriptsize\texttt{[['car', 'slow']]} &Contr. L &0.76 &0.50 &0.40 &0.50 &0.71 \\
\bottomrule
\end{tabular}
}
\caption{Comparing the behavior of FactSim against ROUGE (R-1, R-2, and R-L) and BERTScore (BS) when comparing a reference claim ("The car is fast") against sentences that entail or contradict it. We consider negations (N), synonyms (P), and the use of longer expressions (L).}
\label{tab:metrics_comparison}
\end{table*}
}

{
\renewcommand{\arraystretch}{1.1}
\begin{table*}[ht]
\footnotesize
\centering
\resizebox{\linewidth}{!}
{
\begin{tabular}{lcccccccccc}\toprule
\multirow{2}{*}{\textbf{metric}} &\multicolumn{2}{c}{\textbf{Asp.Rel.}} &\multicolumn{2}{c}{\textbf{Sel.Coh.}} &\multicolumn{2}{c}{\textbf{Sen.Con.}} &\multicolumn{2}{c}{\textbf{Read.}} &\multicolumn{2}{c}{\textbf{avg. rank}} \\\cmidrule{2-11}
&\textbf{sys} &\textbf{sum} &\textbf{sys} &\textbf{sum} &\textbf{sys} &\textbf{sum} &\textbf{sys} &\textbf{sum} &\textbf{sys} &\textbf{sum} \\\midrule
BARTScore$_{rev\rightarrow hyp}$ &\textbf{\textcolor{red}{0.65}}$^{**}$ &\textbf{0.22} &\textbf{\textcolor{red}{0.76}}$^{**}$ &\textbf{\textcolor{red}{0.29}} &\textbf{\textcolor{red}{0.77}}$^{**}$ &\textbf{\textcolor{red}{0.34}} &\textbf{0.46}$^{*}$ &\textbf{0.33} &1.5 &2.3 \\
BLANC$_{help}$ &\textbf{0.56}$^{**}$ &\textbf{0.17} &0.54$^{**}$ &\textbf{0.16} &\textbf{0.62}$^{**}$ &\textbf{0.24} &0.38 &\textbf{0.17} &5.5 &4.8 \\
BLANC$_{tune}$ &0.49$^{*}$ &0.14 &0.47$^{*}$ &0.10 &0.55$^{**}$ &0.19 &0.31 &0.09 &8.5 &8.3 \\
PPL-\texttt{[PEGASUS]} &-0.08 &-0.06 &-0.01 &-0.07 &0.02 &-0.03 &-0.22 &-0.07 &14.8 &15.0 \\
SUPERT &0.54$^{**}$ &\textbf{0.17} &\textbf{0.56}$^{**}$ &\textbf{0.14} &\textbf{0.60}$^{**}$ &0.18 &\textbf{0.40}$^{*}$ &0.12 &6.3 &6.8 \\
QAFactEval &0.45$^{*}$ &0.08 &0.47$^{*}$ &0.09 &0.51$^{*}$ &\textbf{0.21} &0.27 &0.13 &9.8 &8.3 \\
SummaQA$_{fscore}$ &\textbf{0.56}$^{**}$ &0.10 &\textbf{0.58}$^{**}$ &0.09 &\textbf{0.66}$^{**}$ &0.17 &0.38 &0.10 &4.5 &9.0 \\
SummaQA$_{conf}$ &\textbf{0.58}$^{**}$ &0.10 &\textbf{0.60}$^{**}$ &0.12 &\textbf{0.69}$^{**}$ &0.14 &\textbf{0.44}$^{*}$ &\textbf{0.15} &2.8 &8.3 \\
SummaC$_{snt}$ &0.19 &0.00 &0.16 &0.01 &0.22 &0.12 &-0.04 &0.01 &12.8 &12.8 \\
SummaC$_{doc}$ &0.30 &0.06 &0.27 &0.01 &0.40 &0.17 &0.11 &0.05 &11.5 &11.3 \\
PPL-\texttt{[GPT-2]} &-0.10 &-0.14 &-0.08 &-0.15 &0.00 &-0.05 &-0.24 &-0.16 &16.0 &16.0 \\
G-Eval-\texttt{[text-ada-001]} &-0.01 &-0.01 &-0.05 &-0.02 &0.22 &0.08 &0.27 &0.08 &13.3 &13.0 \\
G-Eval-\texttt{[text-ada-001]-n} &0.05 &0.01 &0.12 &\textbf{0.14} &0.40 &0.07 &0.29 &0.06 &11.8 &11.0 \\
G-Eval-\texttt{[gpt-3.5-turbo]} &0.45$^{*}$ &\textbf{0.23} &\textbf{0.56}$^{**}$ &\textbf{0.26} &0.55$^{**}$ &\textbf{\textcolor{red}{0.34}} &\textbf{0.53}$^{**}$ &\textbf{0.36} &6.3 &2.0 \\
ChatGPT-\texttt{[gpt-3.5-turbo]} &\textbf{0.56}$^{**}$ &\textbf{\textcolor{red}{0.30}} &\textbf{0.62}$^{**}$ &\textbf{0.25} &0.56$^{**}$ &\textbf{0.33} &\textbf{\textcolor{red}{0.62}}$^{**}$ &\textbf{\textcolor{red}{0.42}} &3.3 &2.0 \\
FactSim (ours) &\textbf{0.56}$^{**}$ &\textbf{0.26} &\textbf{0.58}$^{**}$ &\textbf{0.21} &\textbf{0.62}$^{**}$ &\textbf{0.27} &\textbf{0.42}$^{*}$ &\textbf{0.16} &4.3 &3.8 \\
\bottomrule
\end{tabular}
}
\caption{Kendall's $\tau$ correlations at system- and summary level between automatic metrics and human judgments along four different dimensions, for metrics that do not require a reference summary for evaluation (those marked with $\blacktriangledown$ in \cite{shen2023opinsummeval}). Results marked with $*$ and $**$ denote system-level scores with p-values of $\leq 0.01$ and $\leq 0.005$, respectively. The last two columns show the average ranking across the four evaluation dimensions. Top-6 scores are in boldface while the best in each category is highlighted in red.}
\label{tab:kendall_correlations}
\end{table*}
}

\subsection{Experimental setup}

We used gpt-4 to extract tuples (prompt in Section \ref{method}.2) and temperature $0.0$. Tuples were converted to a single embedding using the \texttt{distiluse-base-multilingual-cased} available at HuggingFace \cite{wolf2019huggingface}. For similarity scores in Eqs.~\eqref{eq:fV} and \eqref{eq:fN}, we use the cosine similarity constrained to the range $[0,1]$, \ie $sim(a,b)=\max\{0, \frac{a^Tb}{\|a\|\|b\|}\}$. ROUGE scores were calculated using Python library \texttt{[rouge-score 0.1.2]} and BertScore using library \texttt{[bert-score 0.3.13]} with model \texttt{bert-base-uncased}. Summaries generated in this work (such as in Section \ref{sec:results}.3) used the prompt in Appendix \ref{apx:summatization_promp} and temperature $0.0$.

\section{Results}\label{sec:results}

In the following subsections we address different aspects of our approach: its use as a factuality score, as a metric for opinion summarization, and its explainable nature.

\subsection{FactSim as a factuality score}

In the first set of experiments, we compare our metric with others like ROUGE (R-1, R-2, and R-L) and BERTScore (BS) in a simple but illustrative example. Table~\ref{tab:metrics_comparison} shows the scores obtained by the different metrics when comparing the expressions in the first column against a reference claim (marked with an asterisk in the table). The first group shows scores for claims that entail the same information as the reference (Ent.), while the second shows scores for sentences that contradict the reference claim (Contr.). For each group, we consider negations (N), synonyms (P), and the use of longer expressions (L). We show fact-tuples extracted by our method in each case. From the table, we see that ROUGE-based metrics exhibit an overall unstable behavior both against rephrasings that keep the semantic intent as when it contradicts it. Moreover, we see that scores for contradicting claims might be even higher compared to entailed ones, especially when dealing with negations. A similar behavior can be observed in BERTScore, although with more stable scores within each group. Our metric, on the contrary, provides consistently higher scores for entailments than for contradictions.

Figure~\ref{fig:metrics_comparison_matrices} shows similarity matrices obtained by ROUGE-1, BERTScore, and FactSim, for the examples in Table~\ref{tab:toy_examples}. In these examples, sentences 1-3 and 2-4 entail similar information. Both ROUGE-1 and BERTScore mistakenly attribute higher scores to pairs 1-2 and 3-4, impacted by the negation and the structure of the sentence. Our metric, on the other hand, compares only the tuples, which converts the description to a 1-word paraphrased descriptor.

\subsection{FactSim as a summarization score}

In this section, we follow \cite{shen2023opinsummeval} and evaluate how well $FactSim$ aligns with human judgments along the following dimensions: aspect relevance (Asp.Rel.), self-coherence (Sel.Coh.), sentiment consistency (Sen.Con.), and readability (Read.). Table~\ref{tab:kendall_correlations} shows system- (sys) and summary-level (sum) correlations using the Kendall's $\tau$ coefficient. We only consider metrics that do not require a reference summary for evaluation, as they were shown to outperform those that need it \cite{shen2023opinsummeval}. The last two columns in the table show the average ranking across the four dimensions for both system- and summary-level scores. Top-6 models are highlighted in bold while the best ones are in red. Results marked with $*$ and $**$ denote system-level correlations with p-values of $\leq 0.01$ and $\leq 0.005$, respectively. As noted in \cite{shen2023opinsummeval}, BARTScore$_{rev\rightarrow hyp}$ appears as the best performant model across most categories. Note, however, that this model uses a backbone model fine-tuned in the summarization dataset CNNDM \cite{hermann2015teaching}. In our case, besides not requiring any fine-tuning, we limit our backbone to the task of fact-tuples extraction only. 
Despite its simplicity, our approach consistently ranks among the top-performing ones. Among the four evaluation dimensions, our model performs best on aspect relevance (Asp.Rel.), as it is the task that better aligns with the rationales behind our approach. 

\subsection{Fact-tuple analysis for explainable summaries}

\begin{figure}[!ht]
  \centering
  \includegraphics[width=\linewidth]{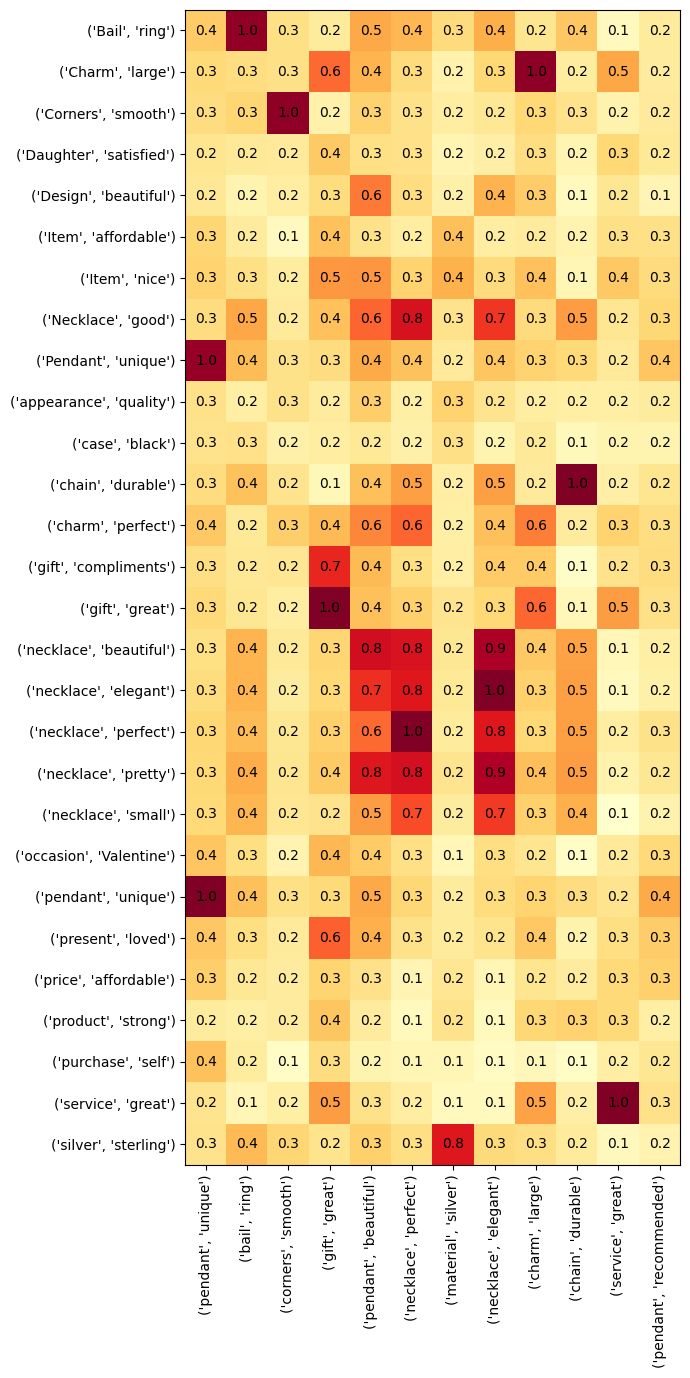}
  \caption{Fact similarity for a sample product. The X-axis contains facts from the summary and the Y-axis facts from source reviews. Notice that \texttt{(price, affordable)} and \texttt{product, strong)} are claims not covered by the LLM. The LLL, however, included facts that are not present in the original reviews, such as \texttt{(pendant, recommended)}, which is a conclusion created by \texttt{gpt-3.5-turbo} based on overall sentiment of the reviews.}
  \label{fig:sim-matrix-example}
\end{figure}

Working with fact-tuples allows for a post hoc analysis of the main aspects highlighted in the summary. Figure~\ref{fig:sim-matrix-example} illustrates the similarity scores between the fact-tuples extracted from the pool of source reviews and those extracted from the summary, for a sample taken from the gold standard in \cite{bravzinskas2019unsupervised} and with the summary generated by \texttt{gpt-3.5-turbo}. Rows in this matrix represent tuples from the source reviews, while columns encode tuples from the summary. Notice that some reviews include aspects that were not covered by the summary, such as \texttt{(price, affordable)} or \texttt{(product, strong)}, while some facts in the summary correspond to overall feelings towards the product that were included by the model, \eg \texttt{(pendant, recommended)} and which are not present in the original reviews. The prompt we used for generating the summaries can be found in Appendix~\ref{apx:summatization_promp}. The set of input reviews and generated summary can be found in Appendix~\ref{apx:fact_similarities}.

\section{Conclusion}
\label{conclusion}

We present in this work a novel metric called $FactSim$ that measures the quality of opinion summaries by comparing factual assertions present both in the source reviews as well as in the summary outputted by a black box model. We argue that a good opinion summary should encompass three key aspects: (i) be fluent and readable, (ii) contain the most cited facts in the source reviews and (iii) do not contain facts that are less relevant or not cited in the source reviews. The first aspect has advanced a lot over the last few years, especially with LLM summarization. The second and third aspects are still an open question since they are either not considered by metrics that are calculated versus a reference summary or are not transparent in neural-based metrics. The proposed metrics advance in this direction, aiming to provide a good measurement in terms of fact coverage and consistency. We found that it has a high correlation to human judgment, especially on aspect-related dimensions, ranking among top state-of-the-art metrics. Evaluation of the fluency and readability dimensions remains as a research direction in future works.

\section{Limitations}
\label{limitations}

\subsection{Annotated dataset}

Annotated datasets are usually scored qualitatively. While the proposed metric obtained a good correlation with dimensions such as aspect relevance, self-coherence, and sentiment consistency, we still believe there is a need for datasets annotated with fact count, which is the basis of our metric. Also, we believe our metric benefits from larger datasets, were claims are constantly repeated throughout different reviews. This characteristic is not present in the benchmark dataset and can be overlooked by current metrics that were compared to ours. Although not in the scope of this work, annotating large datasets with quantitative data may help develop this research field.

\subsection{Information extraction}

We propose a method to evaluate the factual consistency of opinion summarization while providing some explanation of how the claims in the original texts are considered in the summaries. Still, we leveraged the ability of large language models in the task of information extraction. This task alone could lead to inconsistencies that are not being considered, although they tend to have the same effect when comparing the models and, thus, not changing the final result. It is common for reviews to be written in an informal and not structured manner, such as in "my 3 year old fit into these perfectly." (extracted from \citet{bravzinskas2019unsupervised}). This sentence does not specify the product, neither provide any useful description of the product posing a challenge to the way we extract information, but also to other models. A possible solution could be providing context to the LLM or filtering these reviews out.

\bibliography{custom}

\clearpage

\lstset{
  basicstyle=\ttfamily,
  columns=fullflexible,
  frame=single,
  breaklines=true
}

\appendix

\section{Summarization Prompt}
\label{apx:summatization_promp}

\begin{lstlisting}[basicstyle=\small,language=Python]
"""Your task is to generate a summary review, given a list of customer reviews of a product.
Use the same words the customers use to refer to the product and its characteristics.
The summary review has a limit of 100 words.

Review list:"""
\end{lstlisting}

\section{Facts similarities for a sample product}
\label{apx:fact_similarities}

\bigskip
Reviews for product B0040EIHQQ from the evaluation set of \cite{bravzinskas2019unsupervised}:

\bigskip
\footnotesize
'This pendant is so unique!! The design is beautiful and the bail is a ring instead of the typical bail which gives it a nice touch!! All the corners are smooth and my daughter loves it - looks great on her.I cannot say anything about the chain because used our own chain.:) Satisfied.'

\bigskip
'It look perfect in a womens neck!! great gift, I thought for the price it was going to look cheap, but I was far wrong. It look great.Spect great reward from your woman when you give this to her; D'

\bigskip
'The prettiest sterling silver piece I own now. I get so many compliments on this necklace. I bought it for myself from my hubby for Valentine's Day. Why not? When people ask where I got it, I simply say from my loving hubby. And he is off the hook as to what to get me. win + win.'

\bigskip
'I love hearts and I love 'love':) I do not have any negative feedback, the necklace is perfect and the charm is perfect. I just thought it would have been slightly bigger. Overall, I love my new heart necklace.'

\bigskip
'When I received the package, I was surprised and amazed because the necklace is so elegant, beautiful and the same as the picture shown here. I really love this necklace. It has a unique pendant designed. I will recommend it to someone to order it now...'

\bigskip
'Item is nice. Not a great quality item, but right for the price. Charm was larger than I expected (I expected small and elegant, but it was large and almost costume jewelry like). I think it is a good necklace, just not what I expected.'

\bigskip
'I got this as a present for my GF on Valintines day. She loves it and wears it every day! Its not cheap looking and it hasn't broken yet. The chain hasn't broken either even though it is very thin. Strongly recomend it!'

\bigskip
'Over all service has been great the only problem, I ordered a purple Mickey Mouse case for iPhone 4S they sent a black, n I felt it was to much trouble n such a small item to send back so needless to say its put back in a drawer somewhere'

\bigskip
\normalsize
Summary generated by \texttt{gpt-3.5-turbo}:

\bigskip
\footnotesize
The pendant is unique and beautiful with a ring bail, giving it a nice touch. Smooth corners and loved by daughters. Great gift for women, looks perfect and not cheap at all. Sterling silver piece with many compliments received. Perfect heart necklace, slightly smaller than expected but still loved. Elegant and beautiful necklace, exactly as shown in the picture. Good quality item for the price, charm larger than expected. Not cheap looking, durable chain. Overall great service, minor issue with wrong color sent. Recommended for a unique pendant design.

\normalsize

\end{document}